%% file: 0_main.tex
\documentclass[final]{cvpr}

\usepackage{times}
\usepackage{epsfig}
\usepackage{graphicx}
\usepackage{amsmath}
\usepackage{amssymb}

\usepackage{url}            
\usepackage{booktabs}       
\usepackage{amsfonts}       
\usepackage{nicefrac}       
\usepackage{microtype}      
\usepackage{enumerate}
\usepackage{xcolor}
\usepackage[normalem]{ulem}
\usepackage[caption=false,font=footnotesize]{subfig}
\usepackage{booktabs}
\usepackage{bigstrut}
\usepackage{stackengine}
\usepackage{multirow}

\newcommand{\red}[1] {\textcolor[rgb]{1.0,0.0,0.0}{{#1}}}

\newcommand{\vect}[1]{\mathbf{#1}}

\def\Put(#1,#2)#3{\leavevmode\makebox(0,0){\put(#1,#2){#3}}}

\newcommand{\cv}{\vect{c}}

\newcommand{\xv}{\vect{x}}
\newcommand{\Xv}{\vect{X}}

\newcommand{\uv}{\vect{u}}
\newcommand{\Uv}{\vect{U}}

\newcommand{\Iv}{\vect{I}}
\newcommand{\pv}{\vect{p}}
\newcommand{\R}{\ensuremath{\mathbb{R}}}

\usepackage{amsthm}

\theoremstyle{definition}

\def\ie{{\it i.e.}}
\def\eg{{\it e.g.}}
\def\etc{{\it etc.}}
\def\etal{{\it et~al.}}

\usepackage[pagebackref=true,breaklinks=true,colorlinks,bookmarks=false]{hyperref}

\def\cvprPaperID{4464} 
\def\confYear{CVPR 2021}

\begin{document}

\title{TearingNet: Point Cloud Autoencoder to Learn Topology-Friendly Representations}
\def\namespacing{15pt}
\def\emailspacing{3pt}
\author{%
   Jiahao~Pang\hspace{\namespacing}Duanshun~Li\thanks{Work done while the author was an intern at InterDigital.}\hspace{\namespacing}Dong~Tian\\
  InterDigital, Princeton, NJ, USA\\
  {\tt\small jiahao.pang@interdigital.com,\hspace{\emailspacing}duanshun@ualberta.ca, dong.tian@interdigital.com}
}

\maketitle
\pagestyle{empty}
\thispagestyle{empty}

\begin{abstract}
Topology matters.
Despite the recent success of point cloud processing with geometric deep learning, it remains arduous to capture the complex topologies of point cloud data with a learning model.
Given a point cloud dataset containing objects with various genera, or scenes with multiple objects, we propose an autoencoder, TearingNet, which tackles the challenging task of representing the point clouds using a fixed-length descriptor.
Unlike existing works directly deforming predefined primitives of genus zero (\eg, a 2D square patch) to an object-level point cloud, our TearingNet is characterized by a proposed Tearing network module and a Folding network module interacting with each other iteratively.
Particularly, the Tearing network module learns the point cloud topology explicitly.
By breaking the edges of a primitive graph, it tears the graph into patches or with holes to emulate the topology of a target point cloud, leading to faithful reconstructions.
Experimentation shows the superiority of our proposal in terms of reconstructing point clouds as well as generating more topology-friendly representations than benchmarks.
\end{abstract}


\section{Introduction}\label{sec:intro}
\input{1_intro}

\section{Related Work}\label{sec:related}
\input{2_related}

\section{Topology Update with the TearingNet}\label{sec:tearing}
\input{3_tearing}

\section{The TearingNet Architecture}\label{sec:arch}
\input{4_arch}

\section{Experimentation}\label{sec:results}
\input{5_results}

\vspace{-2pt}
\section{Conclusion}\label{sec:conclude}
\vspace{-5pt}
\input{6_conclude}

{\small
\bibliographystyle{ieee_fullname}

}

\end{document}

%% file: 1_intro.tex
\def\demopcwidth{0.28}
\def\demogridwidth{0.25}
\def\demoinputwidth{0.25}
\def\demomargin{-4pt}
\def\democaption{-3pt}
\begin{figure}[htbp]
  \centering\footnotesize
    \begin{tabular}{c||c|cc}
    \hline
    (a)\,Input &       & \multicolumn{2}{c}{$\textrm{(b)\,2D primitive}\xrightarrow[]{\textrm{Folding}}\textrm{(c)\,3D point cloud}$} \bigstrut\\
    \hline
    \multirow{2}[2]{*}{
    \hspace{\demomargin}\hspace{\demomargin}
    \begin{minipage}[c]{\demoinputwidth\columnwidth}\center 
        \includegraphics[width=1\columnwidth]{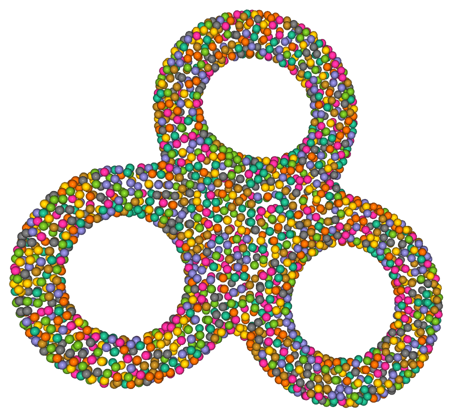}
    \end{minipage}
    \hspace{\demomargin}\hspace{\demomargin}
    } & \multicolumn{1}{c|}{\multirow{2}[2]{*}{\hspace{\demomargin}\rotatebox{90}{$\textrm{(ii)\,After}\xleftarrow[]{\textrm{Tearing}}\textrm{(i)\,Before}$}\hspace{\demomargin}}} & 
    \begin{minipage}[c]{\demogridwidth\columnwidth}\center 
        \vspace{1pt}\includegraphics[width=1\columnwidth]{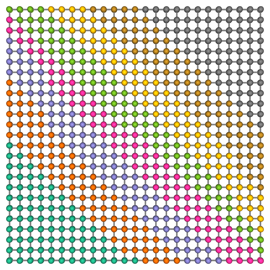}
    \end{minipage}   
    & 
    \begin{minipage}[c]{\demopcwidth\columnwidth}\center 
        \vspace{1pt}\includegraphics[width=1\columnwidth]{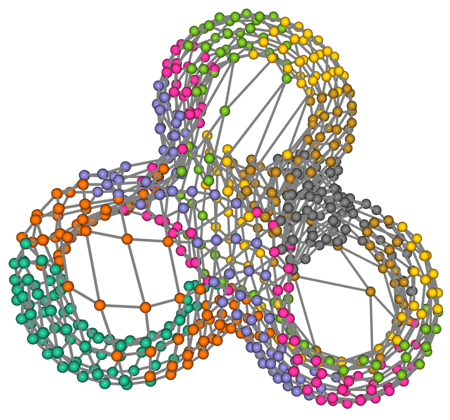}
    \end{minipage}
    \bigstrut[t]\\       &       & 
    \begin{minipage}[c]{\demogridwidth\columnwidth}\center 
        \includegraphics[width=1\columnwidth]{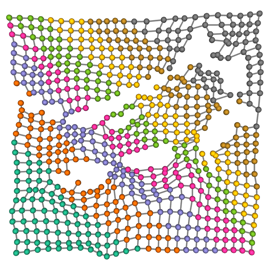}
    \end{minipage}    
    & 
    \begin{minipage}[c]{\demopcwidth\columnwidth}\center 
        \includegraphics[width=1\columnwidth]{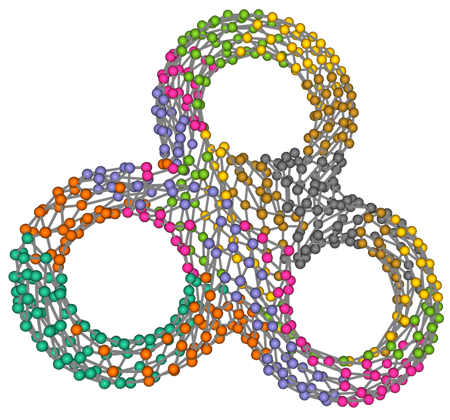}
    \end{minipage} 
    \bigstrut[b]\\
    \hline
    \multirow{2}[2]{*}{
    \hspace{\demomargin}\hspace{\demomargin}
    \begin{minipage}[c]{\demoinputwidth\columnwidth}\center 
        \includegraphics[width=1\columnwidth]{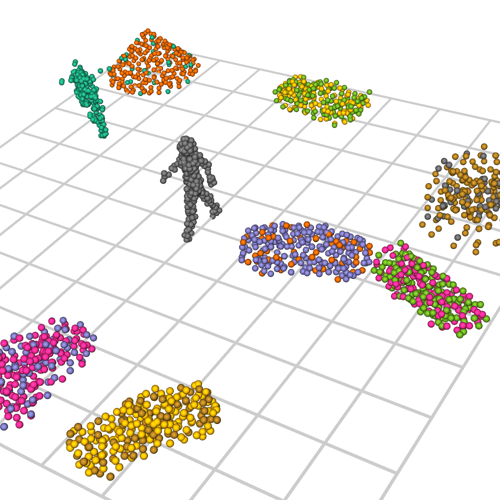}
    \end{minipage}
    \hspace{\demomargin}\hspace{\demomargin}
    } & \multicolumn{1}{c|}{\multirow{2}[2]{*}{\hspace{\demomargin}\rotatebox{90}{$\textrm{(iv)\,After}\xleftarrow[]{\textrm{Tearing}}\textrm{(iii)\,Before}$}\hspace{\demomargin}}} & 
    \begin{minipage}[c]{\demogridwidth\columnwidth}\center 
        \vspace{1pt}\includegraphics[width=1\columnwidth]{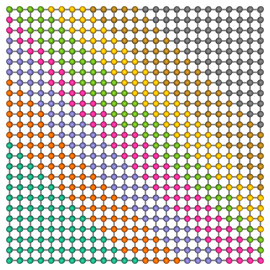}
    \end{minipage}   
    & 
    \begin{minipage}[c]{\demopcwidth\columnwidth}\center 
        \vspace{1pt}\includegraphics[width=1\columnwidth]{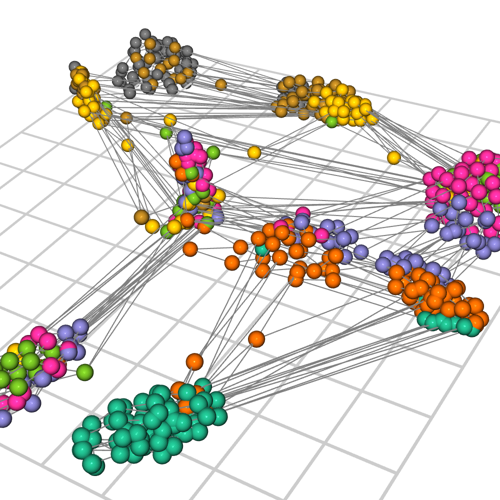}
    \end{minipage}
    \bigstrut[t]\\       &       & 
    \begin{minipage}[c]{\demogridwidth\columnwidth}\center 
        \includegraphics[width=1\columnwidth]{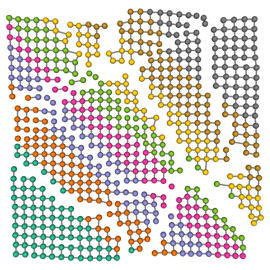}
    \end{minipage}
    & 
    \begin{minipage}[c]{\demopcwidth\columnwidth}\center 
        \includegraphics[width=1\columnwidth]{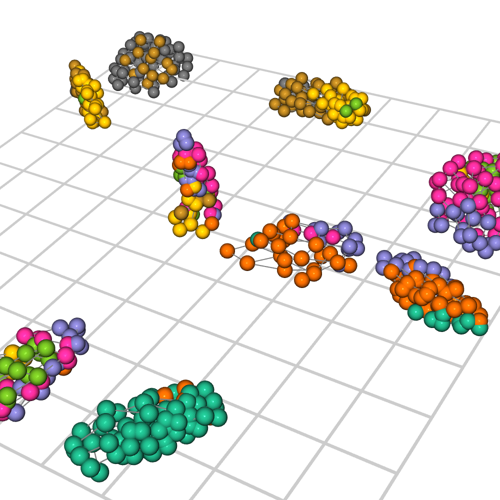}
    \end{minipage}
    \bigstrut[b]\\
    \hline
    \end{tabular}%
    \vspace{5pt}
     \caption{\small TearingNet for point cloud reconstruction. By \emph{Tearing} and \emph{Folding}, we achieve high-quality reconstructions (c-ii) and (c-iv). Edges of the primitive graphs are also drawn in (b) and (c).}
  \label{fig:demo}%
  \vspace{-10pt}
\end{figure}%

Based on a point cloud sampled from the surface of an object (or a scene), humans are able to perceive its underlying shape.
Via properly capturing the \emph{topology} behind the point set, human understanding is robust to variations in scales and viewpoints.
Intuitively, topology reflects how the points are put together to form an object.
Moreover, topology is an intrinsic property of Riemannian manifolds that are usually used to model 3D shapes in geometric learning~\cite{boscaini2016learning, masci2015geodesic}.
Hence, it is important to seek topology-aware representations for point clouds in machine learning.

As an unsupervised learning architecture, \emph{autoencoder} (AE)~\cite{ng2011sparse} is popularly investigated to learn latent representations with unlabeled point clouds.
It tries to approximate an \emph{identity} function that is non-trivially constrained by outputting a compact representation from its encoder network.
The decoder network attempts to reconstruct the point cloud from the compact representation.
The compact representation typically takes the shape of a fixed-length codeword which characterizes geometric properties of point clouds. 
Therefore, it not only preserves the ability for reconstruction~\cite{chen20203d} but is also valuable for high-level tasks such as classification~\cite{yang2018foldingnet, zhao20193d, gao2020graphter}.

With ample topological structures in the real world, unfortunately, it is non-trivial to produce topology-friendly representations that count for object point clouds with holes (\ie, varying genera) or scene point clouds with a varying number of objects (\ie, varying zeroth Betti number~\cite{spanier1989algebraic}).
For example, to represent a multi-object scene, a descriptor not only needs to delineate the shapes of individual objects but also the spatial relationship between them.
In fact, existing works, including LatentGAN~\cite{achlioptas2018learning}, FoldingNet~\cite{yang2018foldingnet}, AtlasNet~\cite{groueix2018papier}, 3D Point Capsule Network~\cite{zhao20193d} and
GraphTER~\cite{gao2020graphter}, all target to reconstruct point clouds with simple topologies, \eg, object point clouds.
Some other works, such as AtlasNetV2~\cite{deprelle2019learning} and DeepSDF~\cite{park2019deepsdf}, even train class-specific networks to alleviate the effects of ample topologies across classes. 





To tackle this challenge, we propose a new autoencoder for point cloud, entitled \emph{TearingNet}.
Similar to FoldingNet~\cite{yang2018foldingnet}, AtlasNet~\cite{groueix2018papier} and 3D Point Capsule Network~\cite{zhao20193d}, \etc, we reconstruct point clouds by deforming genus zero 2D primitives (\eg, a regular 2D patch).
Differently, we novelly \emph{tear} the primitive with holes or into several parts to match its topology to the target point clouds, \eg, Figure~\ref{fig:demo}\red{b-ii} and Figure~\ref{fig:demo}\red{b-iv}.
Our architecture couples a proposed Tearing network and a Folding network, letting them interact with each other.
Especially, by running through a trial folding in the first place, our proposal enables the Tearing network to tear a graph defining on the 2D primitive, letting it count for the empty space in object holes and boundaries.
The torn primitive graph, with desired topology similar to the target point cloud, is fed to the Folding network again.
In this way, the Tearing network and the Folding network run alternatively to decode an accurate reconstruction.

We verified that the TearingNet generates topology-friendly representations.
The superiority of the obtained representations is demonstrated in different down-stream tasks, including shape reconstruction, object counting, and object detection.
We also examine why the learned representations are topology-aware by analyzing the feature space.

The contributions of our work can be summarized below:
\begin{enumerate}[(i)]
\vspace{-5pt}
\item 
We propose a novel autoencoder---TearingNet---to generate topology-friendly representations for complex 3D point clouds.
Our TearingNet includes a Folding network (F-Net) and a Tearing network (T-Net) ``talking'' to each other by a feedback loop, which gradually improves the reconstructions.
\vspace{-5pt}
\item 
Our TearingNet endeavors to explicitly learn the target point cloud topology.
By tearing a local graph built upon the 2D primitive, we update its topology to match with the ground-truth point cloud.
With the notion of graph tearing, our proposal is generalized as a Graph-Conditioned AutoEncoder (GCAE) which discovers and utilizes topology iteratively.
\vspace{-5pt}
\item 
We unroll the proposed TearingNet then apply its produced representations to several tasks that are sensitive to topology, \eg, reconstruction, and object counting, which verify our superiority.
We further analyze the structure of the feature space to understand how the topology-friendliness is achieved.
\end{enumerate}

Our paper is organized as follows. 
Section~\ref{sec:related} reviews related work.
In Section~\ref{sec:tearing}, we present the methodology of the TearingNet under the notion of graph tearing.
Section~\ref{sec:arch} unrolls the TearingNet architecture then illustrates its individual components.
Experimentation is presented in Section~\ref{sec:results} and conclusions are provided in Section~\ref{sec:conclude}.

%% file: 2_related.tex
Recently, geometric deep learning has shown great potential in various point cloud applications~\cite{guo2020deep}.
Compared to deep learning on regularly structured data such as image and video, point cloud learning is, however, more challenging as the points are unorganized and irregularly sampled over the object/scene surface.

\textbf{Non-native learning of point cloud}:
Conventionally, point clouds are preprocessed, \eg, either voxelized~\cite{maturana2015voxnet, ioannidou2017deep} or projected into multiview images~\cite{su2015multi}, so as to carry over deep learning frameworks justified in the image domain. 
After format conversion, for example, the conventional convolutional neural network (CNN) could be applied on 3D voxels or 2D pixels~\cite{choy20163d, roveri2018network}. 
Obviously, voxelization exhibits a tradeoff between accuracy and data volume; while multiview projection is a balance between accuracy/occlusion and data volume. 
Such compromises occur even before the data is fed into deep neural networks. 
Octree-like approaches~\cite{wang2018adaptive} demonstrate limited adaptivity on such tradeoffs.
Fortunately, emerging techniques for native learning on point clouds relieve the frustration from the front.

\textbf{Point cloud encoders}:
In \cite{qi2017pointnet}, Qi \etal propose \emph{PointNet}, which directly operates on input points and generates a latent codeword depicting the object shape.
The latent code is point permutation invariant through a pooling operation.
Once equipped with object-level or part-level labels, PointNet could serve for supervised tasks like classification or segmentation.
Qi~\etal~\cite{qi2017pointnet} also show that, PointNet is a \emph{universal approximator} of any set functions.
In other words, it is highly flexible network whose behavior heavily depends on the overall network design. 
PointNet++~\cite{qi2017pointnet++} recursively applies PointNet in a hierarchical manner so as to capture local structures and enhance the ability to recognize fine-grained patterns.
With similar motivations, PointCNN~\cite{li2018pointcnn} utilizes a hierarchical convolution and Dynamic Graph CNN (DGCNN)~\cite{wang2019dynamic} employs an edge-convolution over graphs.
In brief, advanced feature extractors for point clouds often exploit local topology information.

\textbf{Point cloud decoders}:
As opposed to the advanced feature extractors, designs of current point cloud generators, \eg, the generator in a Generative Adversarial Network (GAN)~\cite{goodfellow2014generative} and the decoder in an autoencoder (AE)~\cite{ng2011sparse}, appear to be more preliminary without taking the advantage of topology.
For example, topology is never counted by the fully-connected decoder of LatentGAN~\cite{achlioptas2018learning}.
Exploiting the fact that point clouds are sampled from 2D surfaces/manifolds, the pioneering works FoldingNet~\cite{yang2018foldingnet} and AtlasNet~\cite{groueix2018papier} propose to 
fold/deform 2D primitives to reconstruct 3D point clouds.
They for the first time embed topology explicitly with 2D patches, such as square patches, of genus zero in their decoders. 

FoldingNet adopts a PointNet-like~\cite{qi2017pointnet} encoder to produce latent representations.
Like the PointNet encoder, FoldingNet decoder is a point-wise network shared among points.
Given a predefined 2D primitive (\eg, 2D square patch or a sphere), the FoldingNet decoder takes a 2D coordinate from the primitive and latent codeword as input, then maps the 2D point to a 3D coordinate.
The set of 3D points mapped by the FoldingNet decoder constitute the reconstructed point cloud.
Unfortunately, by a direct mapping from the regular 2D samples to 3D points, FoldingNet fails to represent point clouds with complex topologies even if the network is scaled up~\cite{yang2018foldingnet}.
AtlasNet~\cite{groueix2018papier} and AtlasNetV2~\cite{deprelle2019learning} naively duplicate primitive-decoder pairs to comply with different shapes; while 3D Point Capsule Network~\cite{zhao20193d} learns multiple latent capsules to describe individual object parts.
However, these approaches do not scale well for point clouds with complex topologies (Section~\ref{ssec:comp}).
In~\cite{chen2019deep}, a fully-connected graph is advanced as a companion to the FoldingNet decoder aiming to approximate point cloud topology by a graph.
Its main weakness is in the misaligned topology from graphs to point clouds as it allows connections between distant point pairs.

Recent works, such as DeepSDF~\cite{park2019deepsdf} and DISN~\cite{xu2019disn}, propose to use signed distance functions (SDF) as implicit representations of 3D shapes.
However, learning of such representations requires additional knowledge of object surface that is difficult to acquire in practice.
For instance, the widely used LiDAR sensors only provide \emph{sparse} and \emph{incomplete} point clouds \cite{geiger2012we}.

Motivated by the limitations in the related work, we propose an autoencoder---\emph{TearingNet}.
Intuitively, our proposal ``tears'' the 2D primitive into pieces or with holes so as to align its topology to the target 3D point clouds.
It effectively drives the latent representation to be aware of the topology of the point clouds.
To the best of our knowledge, TearingNet is the \emph{first} autoencoder that is able to use a fixed-length latent representation to reconstruct multi-object point clouds.

%% file: 3_tearing.tex

\subsection{Overview}
\begin{figure}
  \centering
  \includegraphics[width=0.9\columnwidth]{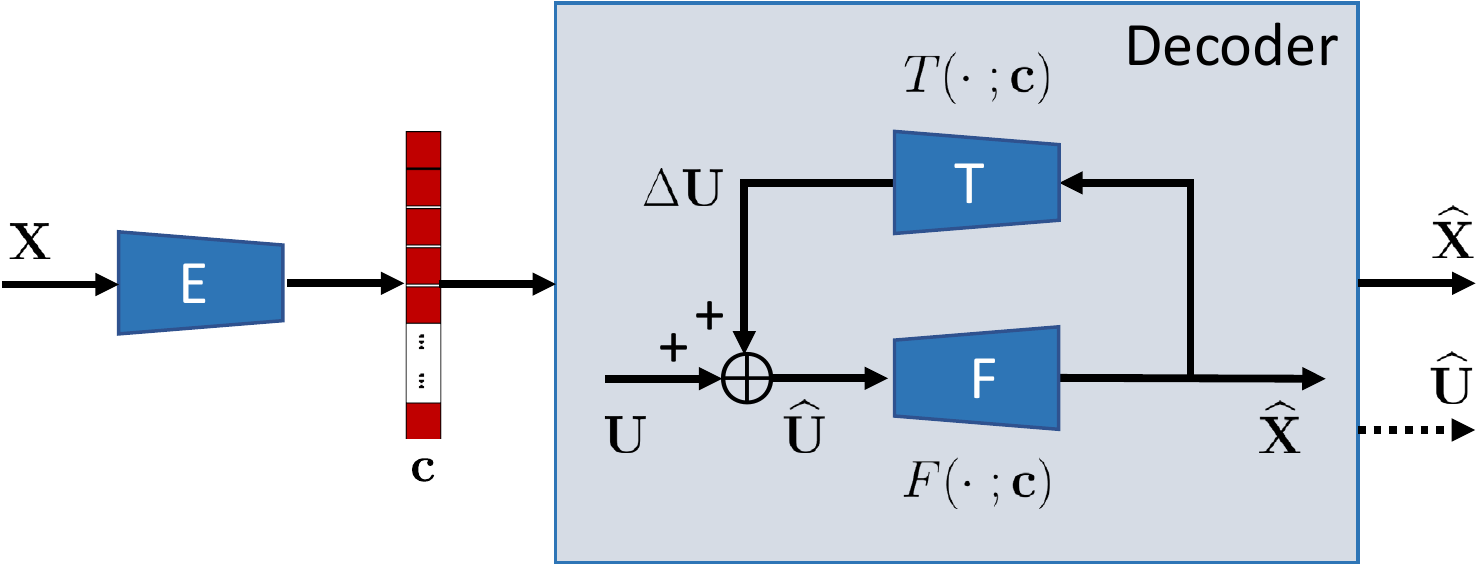}
  \caption{\small Block diagram of the TearingNet/GCAE, featured by the interaction between the F-Net and the T-Net. ``E'', ``F'', ``T'' represent the E-Net, F-Net and T-Net, respectively.}
  \label{fig:gcae}
\end{figure}

A block diagram of the TearingNet is shown in Figure~\ref{fig:gcae}.
The Encoder network (referred to as E-Net, ``E'' in Figure~\ref{fig:gcae}) first generates a latent representation $\cv$ from an input 3D point cloud. 
It is then passed to the TearingNet decoder, which consists of two sub-networks.
On top of the FoldingNet~\cite{yang2018foldingnet} decoder, referred to as \emph{Folding network} (F-Net) hereinafter, a novel \emph{Tearing network} (T-Net) is proposed to couple with the F-Net by a feedback loop.
Given a codeword $\cv$, the TearingNet decoder runs the F-Net and the T-Net (``F'' and ``T'' in Figure~\ref{fig:gcae}) iteratively.
In a nutshell, the F-Net takes as input a certain topology (represented by ${\widehat{\Uv}}$) and ``embeds'' it to a 3D point cloud $\widehat{\Xv}$, then the T-Net considers $\widehat{\Xv}$ to ``correct'' the topology with a feedback connection.

For an input 3D point cloud ${\Xv}=\{{\xv}_i\}_{i=1}^{n}$ composed of $n$ points ${\xv}_i = (x_i, y_i, z_i)$, the encoder first generates a vector ${\cv}\in\R^d$.
As an auxiliary input, a 2D point set $\Uv=\{\uv_i\}_{i=1}^m$ samples $m$ points ${\uv}_i = (u_i, v_i)$ on a 2D region.
Similar to FoldingNet~\cite{yang2018foldingnet}, it contains regularly sampled 2D-grid locations in the square region $[-1,1]\times[-1,1]$ (the points in Figure~\ref{fig:demo}\red{b-i} and Figure~\ref{fig:demo}\red{b-iii}).
This point set $\Uv$ brings in a \emph{primitive} shape for reconstruction, which embodies a \emph{genus zero} topology.
For convenience, the 2D point set is also referred to as \emph{2D grid} in the sequel.
With the 2D grid $\widehat{\Uv}$ (or $\Uv$ at the first iteration), the Folding network maps each of its 2D point $\widehat{\uv}_i$ to a 3D coordinate $\widehat{\xv}_i$, aiming at reconstructing a 3D point cloud $\widehat{\Xv}$ following the topology of $\widehat{\Uv}$.
The Tearing network especially takes $\widehat{\Xv}$ and modifies each 2D point in $\widehat{\Uv}$ individually, leading to a new 2D grid representing an updated topology (\eg, Figure~\ref{fig:demo}\red{b-ii} and Figure~\ref{fig:demo}\red{b-iv}).
Hence, the Folding network and the Tearing network interact with each other for high-quality reconstructions.

\subsection{Tearing as Breaking Graph Edges}\label{ssec:graph_tearing}
With the Tearing network that ``stretches'' the 2D grid, the overall TearingNet still admits a continuous mapping from the 2D space to 3D.
It appears to be \emph{conflict} with the notion of tearing, which implies introducing discontinuities.
We fill this gap by viewing tearing as collectively breaking graph edges that connect neighboring points on the 2D grid.

\textbf{2D grid as a local graph}:
The 2D grid $\Uv$ in our work (as well as FoldingNet~\cite{yang2018foldingnet} and AtlasNet~\cite{groueix2018papier}) essentially approximates a simple Riemannian manifold---a genus zero square patch, denoted as $\mathcal{M}$---with $m$ points regularly sampled on it.
From \cite{hein2007graph,ting2010analysis}, \etc, to represent/approximate a Riemannian manifold (in the continuous domain) with its sampled points (in the discrete domain) means to construct a \emph{local graph} connecting the nearby points.
In other words, the set $\Uv$ implies a primitive graph (denoted as $\mathcal{G}$) associated with the manifold $\mathcal{M}$. 
This graph $\mathcal{G}$, according to \cite{hein2007graph}, have $m$ vertices, with each represents one point in $\Uv$.
For any two points $\uv_i$ and $\uv_j$ from $\Uv$, the graph weight between them is given by a truncated Gaussian kernel:
\begin{equation}\label{eq:graph}
w_{ij} = \left\{ {\begin{array}{*{20}{l}}
{\exp \left( { - \frac{\displaystyle{{{\left\| {{{\bf{u}}_i} - {{\bf{u}}_j}} \right\|}_2^2}}}{\displaystyle{2\epsilon{^2}}}} \right)} & {\mbox{if} ~~ {\left\| {{{\bf{u}}_i} - {{\bf{u}}_j}} \right\|}_2 \le r,}\\
0&{{\rm{otherwise,}}}
\end{array}} \right.
\end{equation}
%
where $\epsilon>0$ is a parameter controlling the sensitivity of the graph weight, and $r>0$ is a threshold.
Hence, the primitive graph $\mathcal{G}$ is an $r$-neighborhood graph, \ie, there is no edge between two points with a distance greater than $r$.

\textbf{Graph tearing}:
Suppose the 2D grid $\mathbf{U}$ has $N^2$ points with a dimension $N\times N$.
According to \cite{hein2006uniform,hein2007graph}, as $r$ in Eq.~\eqref{eq:graph} becomes smaller and $N$ becomes larger (\ie, 2D grid $\mathbf{U}$ sampling $\cal M$ gets denser), the graph $\mathcal{G}$ better approximates $\mathcal{M}$.
Given a certain dimension $N$, we also let $r$ be small, which takes a value just equal to or slightly larger than $2/N$---the horizontal/vertical spacing of neighboring points.\footnote{An even smaller $r$ results in the trivial case which no graph edges exist.}
Then from Eq.~\eqref{eq:graph}, each point $\uv_i$ has four edges connecting to its neighbors---the top, left, bottom, and right points---on the 2D grid $\Uv$.
Hence, before feeding to the Tearing network, $\mathcal{G}$ defaults back to a simple \emph{2D grid graph}, \eg, Figure\,\ref{fig:demo}\red{b-i}.

Running the Tearing network updates the 2D grid $\Uv$ as well as the graph.
Particularly, a 2D point ${\bf u}_i$ is moved to another location $\widehat{\uv}_i$.
Thus, for any two neighboring points in the 2D grid that are pulled apart by the Tearing network, the graph edge between them is naturally \emph{broken}, resulting in an updated graph (denoted at $\widehat{\mathcal{G}}$).
In our proposal, the breakings of all the edges collectively achieve \emph{tearing in the graph domain} and update the underlying topology: 
\begin{enumerate}[(i)]
\item When localized edges within the 2D grid are removed by the Tearing network, a gap/seam is introduced to the graph topology. This is to reconstruct a target point cloud with holes (\ie, its genus number $g>0$), see Figure\,\ref{fig:demo}\red{c-ii}.
\item When all the edges connecting two groups of points in the 2D grid are removed by the Tearing network, the graph is \emph{torn} apart into disconnected sub-graphs. This benefits the reconstruction when the two groups correspond to two distinct objects. Hence, a point cloud with multiple objects (\ie, its zeroth Betti number $b_0>1$) is reconstructed, see Figure\,\ref{fig:demo}\red{c-iv}.
\end{enumerate}
Note that in practice, both of these two cases may happen on the same point cloud.

\begin{figure}
  \centering \scriptsize 
  \subfloat{\includegraphics[width=.25\columnwidth]{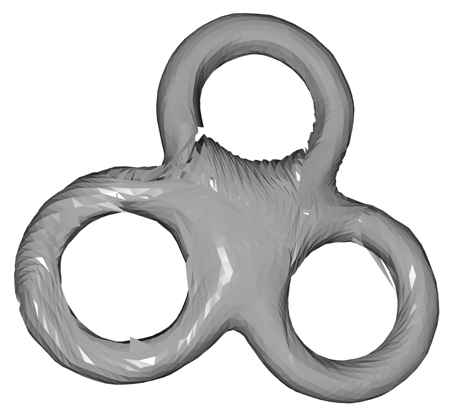}}\hspace{10pt}
  \subfloat{\includegraphics[width=.25\columnwidth]{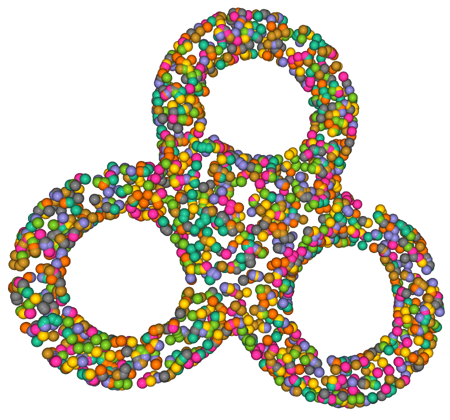}}\hspace{10pt}
  \subfloat{\includegraphics[width=.25\columnwidth]{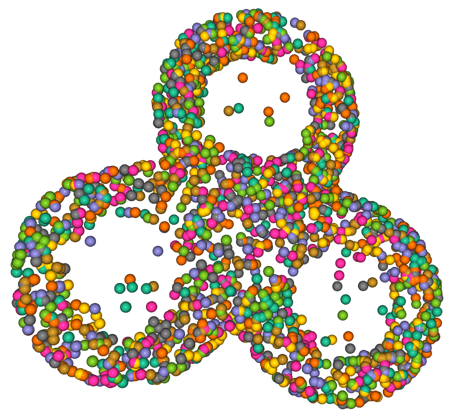}}\\\vspace{-10pt}
  \setcounter{subfigure}{0}
  \subfloat[Induced mesh]{\includegraphics[width=.25\columnwidth]{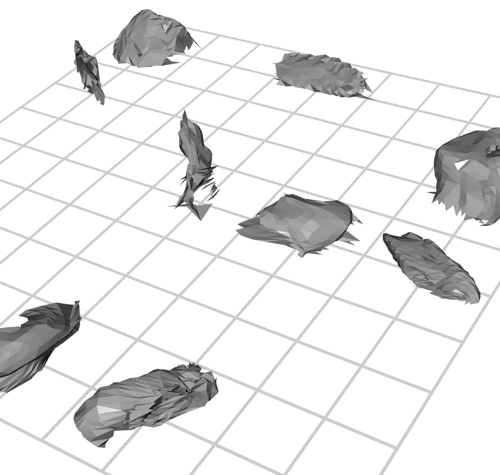}\label{fig:mesh_a}}\hspace{10pt}
  \subfloat[TearingNet]{\includegraphics[width=.25\columnwidth]{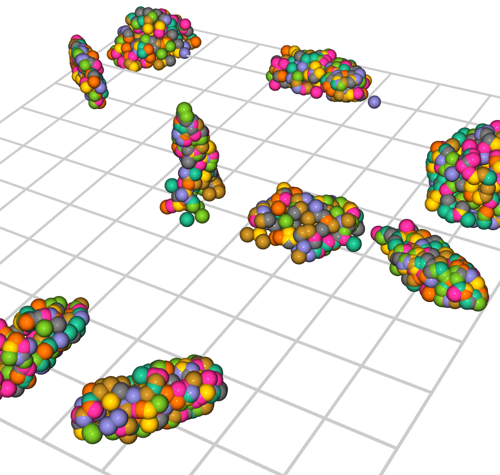}\label{fig:mesh_b}}\hspace{10pt}
  \subfloat[FoldingNet]{\includegraphics[width=.25\columnwidth]{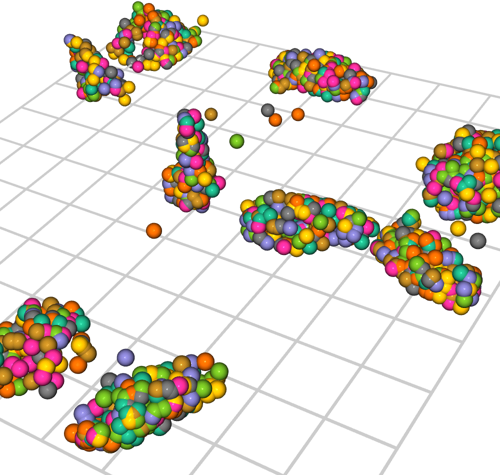}\label{fig:mesh_c}}
  \vspace{3pt}
  \caption{The torn graphs induce 3D meshes (a) and bring better resampled point clouds (b) than that of the FoldingNet (c). Each row in this figure is associated to an example in Figure~\ref{fig:demo}.}
  \label{fig:mesh}
\end{figure}

\textbf{Torn graph as a free mesh}:
The torn graph $\widehat{\mathcal{G}}$---as a side output---naturally represents a mesh over the reconstructed point cloud.
In fact, each elementary square on the 2D grid corresponds to a quadrilateral face of the 3D mesh, where a face is pruned if it has any edges removed by the Tearing network.
Then the remaining faces together constitute a 3D mesh. See Figure~\ref{fig:mesh_a} for an example.

The torn graph also enables us to resample the input 3D point cloud by resampling 2D points in manifold $\mathcal{M}$ while avoiding the ``ghost'' 3D points between different objects or within object holes.
This is achieved by removing outliers sampled on the pruned faces (Figure~\ref{fig:mesh_b}).
In contrast, it is inevitable for the simple FoldingNet to introduce undesired points in the resampled point cloud (Figure~\ref{fig:mesh_c}).

\subsection{Graph-Conditioned AutoEncoder}
The insights of graph tearing motivate us to generalize the architecture of Figure\,\ref{fig:gcae} and call it a \emph{Graph-Conditioned AutoEncoder} (GCAE), which we believe useful for processing data where topology matters, \eg, image, video, or any graph signals.
It promotes an explicit way to \emph{discover} and \emph{utilize} topology within an autoencoder.
Particularly, it is equipped with a graph $\widehat{\mathcal{G}}$ whose topology evolves by iterating F-Net and T-Net.
F-Net \emph{embeds} the graph to a reconstruction; while T-Net attempts to \emph{decodes} a graph (in a residual form) from a reconstruction, which may tear a graph into patches or with holes to achieve desired reconstructions.

%% file: 4_arch.tex
In this section, we unroll the TearingNet (Figure~\ref{fig:gcae}) then elaborate on its components, especially the Tearing network.

\begin{figure*}[htbp]
  \centering
  \includegraphics[width=1.95\columnwidth]{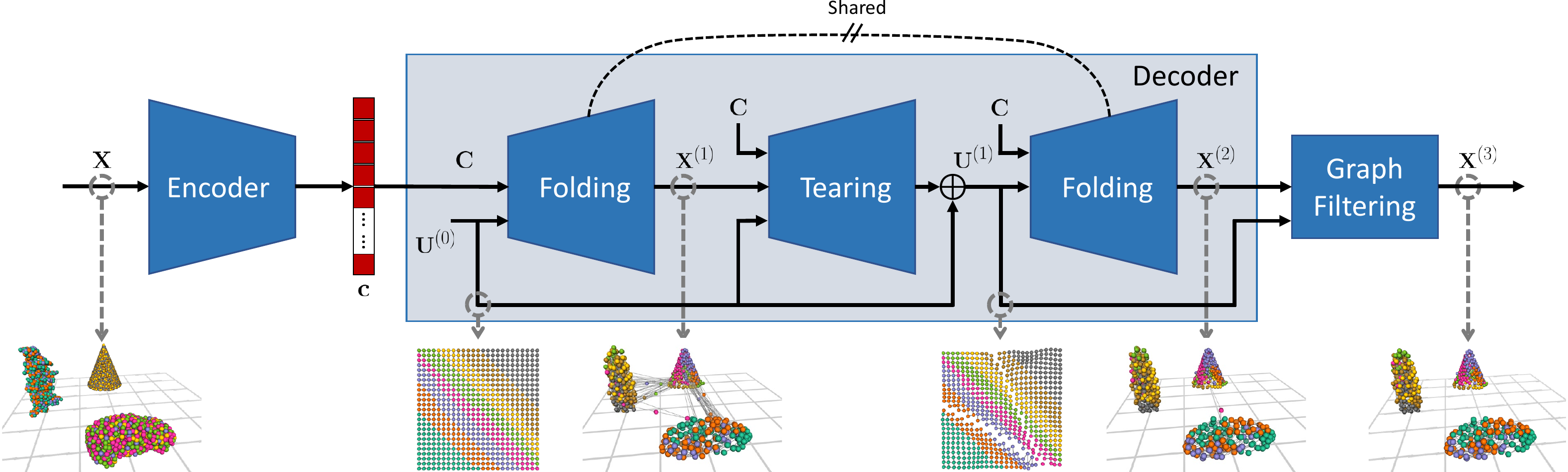}
  \caption{Block diagram of an unrolled TearingNet which has a T-Net wedged in-between two iterations of F-Net.}
  \label{fig:overall}
  \vspace{-5pt}
\end{figure*}

\subsection{Unrolling the TearingNet}
As a concrete example, Figure~\ref{fig:overall} shows an \emph{unrolled} version of the TearingNet where a T-Net is wedged into two iterations of F-Net.
We adopt the PointNet architecture~\cite{qi2017pointnet} as our encoder due to its flexibility (as discussed in Section~\ref{sec:related}): it has the potential to generate topology-friendly representations given it is trained properly.
With the latent code $\cv$ (a $512$-dimension vector in our work) and the initial point set $\Uv^{(0)}$ (the regular 2D grid) as inputs, the TearingNet decoder (in Figure~\ref{fig:overall}) runs F-Net, T-Net and F-Net sequentially.
Specifically, an input point $\uv^{(0)}_i\in {\Uv}^{(0)}$ is mapped to a 3D point $\xv^{(2)}_i$:
\begin{equation}\label{eq:tearing}
\begin{split}
    &\xv^{(1)}_i = F\left(\uv^{(0)}_i; \cv\right)\hspace{5pt}\\
    \rightarrow\hspace{5pt}&\uv^{(1)}_i = T\left(\uv^{(0)}_i, \xv^{(1)}_i;\cv\right)+{\uv^{(0)}_i}\hspace{5pt}\\
    \rightarrow\hspace{5pt}&\xv^{(2)}_i = F\left(\uv^{(1)}_i;\cv\right),
\end{split}
\end{equation}
where $F$ and $T$ denote the \emph{point-wise} networks of F-Net and the T-Net, respectively.
We see that, the F-Net first endeavors a trial folding to produce a preliminary 3D point cloud ${\Xv}^{(1)}$.
The T-Net takes ${\Xv}^{(1)}$ and generates the 2D point set ${\Uv}^{(1)}$.
It is then supplied to the second iteration of F-Net for an improved 3D point cloud ${\Xv}^{(2)}$.

With the updated 2D point set ${\Uv}^{(1)}$ and the second Folding network output ${\Xv}^{(2)}$, one may optionally append a \emph{graph filtering} module at the end to further enhance the reconstruction.
Note that all reconstructions ${\Xv}^{(\cdot)}$ contain $m$ points, the same as that of the 2D pointsets ${\Uv}^{(\cdot)}$.
By iterating the F-Net twice (as Figure~\ref{fig:overall}), the TearingNet achieves a good tradeoff between computation and reconstruction quality.
Hence, we focus on this configuration for experimentation.

\subsection{The Tearing Network}
%
As a core ingredient, the Tearing network (or T-Net) is introduced to explicitly learn the topology by tearing the primitive graph, which boosts the reconstruction accuracy, and ultimately enhances the representability of the codeword.
In our design, the Tearing network learns point-wise modifications to the 2D point set ${\Uv}^{(0)}$ and computes ${\Uv}^{(1)}$ with a residual connection~\cite{he2016deep}, see Figure~\ref{fig:overall}. 
The 2D points are then moved around like flocks depending on the topology chart they belong to.

Similar to PointNet~\cite{qi2017pointnet} and the Folding network~\cite{yang2018foldingnet}, the Tearing network consists of shared point-wise MLP layers.
Its architecture, as well as its model scale, are similar to that of the Folding network.
Given the 2D coordinate $\uv_i^{(0)}\in\mathbb{R}^2$, it is first concatenated with the associated 3D coordinate $ \xv_i^{(1)}\in\mathbb{R}^3$ as well as the codeword $\cv\in\mathbb{R}^{512}$, to form a $517$-dimensional vector.
This vector is then fed to the point-wise network $T$ with two stages of MLP layers, and produces a translation vector on the 2D plane.
The detailed architecture of the Tearing network is provided in the supplementary material.

To demonstrate the effectiveness of the Tearing network, we train the whole TearingNet to over-fit the \emph{Torus} dataset introduced in \cite{chen2019deep} which contains 300 torus-shape point clouds with genus number ranging from 1 to 3.
Figure~\ref{fig:demo}\red{c-i} and Figure~\ref{fig:demo}\red{c-ii} show a genus-3 torus before and after the T-Net respectively, where we see that the 2D grid is ``torn'' with holes to accommodate the topology of the input torus.

\subsection{Graph Filtering for Enhancement}
Equipped with the torn graph, we propose an optional graph filter appended at the end of the TearingNet (see Figure~\ref{fig:overall}) to improve the reconstruction point cloud~\cite{chen2019deep}.
We first compute the unnormalized graph Laplacian matrix ${\bf L}\in\mathbb{R}^{m\times m}$ of the graph $\widehat{\cal G}$.
Then we run the following graph filter~\cite{shuman2013emerging} to obtain the final output $\Xv^{(3)}$:
\begin{equation}\label{eq:graph_filter}
    \Xv^{(3)} = (\Iv - \lambda {\bf L})\cdot\Xv^{(2)}
\end{equation}
where the point clouds $\Xv^{(\cdot)}$ are viewed as $m\times3$ matrices and the filtering parameter $\lambda=0.5$.

For better filtering, we incorporate the second folding output $\Xv^{(2)}$ when computing the edge weights---we let $\pv_i=\left[{\uv^{(1)}_i}^{\rm T}\hspace{5pt}{\xv^{(2)}_i}^{\rm T}\right]^{\rm T}$, and compute ${{\left\| {{{\bf{p}}_i} - {{\bf{p}}_j}} \right\|}_2^2}$ instead of ${{\left\| {{{\bf{u}}_i} - {{\bf{u}}_j}} \right\|}_2^2}$ in Eq.~\eqref{eq:graph}.
Additionally, rather than thresholding with distance $r$, we equivalently remove an edge if its weight is too smaller (\ie, less than $10^{-12}$).
We see that our graph filter is a lightweight and differentiable signal processing module for enhancement with little overhead.
Thus, it is included in the end-to-end training of the unrolled TearingNet.

%% file: 5_results.tex
In this section, the training of the TearingNet, alongside with other experimental settings, are first introduced.
We then perform the evaluation on three tasks: reconstruction, object counting, and object detection.
\subsection{Training of the TearingNet}
Instead of training the entire TearingNet directly, we first pre-train the Encoder network (E-Net) and the Folding network (F-Net).
They are trained together under the FoldingNet~\cite{yang2018foldingnet} autoencoder architecture without the Tearing network.
After that, we load the pre-trained E-Net and F-Net then train the overall TearingNet autoencoder as shown in  Figure~\ref{fig:overall}.
This step specifically lets the Tearing network learn to tear the 2D grid/primitive graph and update the topology.
In this step, a smaller learning rate is adopted.
Details of the training strategy are provided in the supplementary material.

Similar to \cite{chen20203d,yang2018foldingnet}, we train the overall TearingNet (and other methods) with the \emph{augmented Chamfer distance}.
Given an original and a reconstructed point clouds being $\Xv$ and $\widehat{\Xv}$, respectively, the augmented Chamfer distance is written as:
\vspace{-5pt}
\begin{equation}\label{eq:chamfer}
\vspace{-3pt}
  d_{\Xv, \widehat{\Xv}}\hspace{-1pt}=\hspace{-1pt}\max\hspace{-1pt}\left \{\hspace{-1pt}\frac{1}{n}\hspace{-3pt}\sum_{\xv \in \Xv}\hspace{-2pt}\min_{ \widehat{\xv} \in  \widehat{\Xv}}  \left\|  \xv\hspace{-1pt} -\hspace{-1pt} \widehat{\xv} \right\|_2\hspace{-1pt},\hspace{-1pt}\frac{1}{m}\hspace{-3pt}\sum_{ \widehat{\xv} \in  \widehat{\Xv}} \hspace{-2pt}\min_{\xv \in \Xv} \left\|  \xv\hspace{-1pt} -\hspace{-1pt} \widehat{\xv} \right\|_2\hspace{-1pt}\right \}.
\end{equation}
Eq.~\eqref{eq:chamfer} essentially gives the Hausdorff distance between two point clouds \cite{bouaziz2016modern}.
As analyzed in \cite{chen20203d}, it is more robust to ill cases compared to the original Chamfer distance \cite{fan2017point}.

\subsection{Experimental Setup}
\textbf{Datasets}:
We verify our work with both single- and multi-object point cloud datasets, with a focus on the latter ones.
Particularly, we adopt ShapeNet~\cite{chang2015shapenet} and Torus~\cite{chen2019deep} datasets to experiment with the single-object scenarios; while we collect objects from off-the-shelf point cloud datasets to synthesize multi-object point cloud datasets. 
To assemble a point cloud with $k$ objects, a $K\times K$ square-shaped ``playground'' with $K^2$ grids is defined to host objects.
Then randomly picked $k \le K^2$ objects are normalized and randomly placed on the grids of the playground.

We first generate multi-object datasets with objects from KITTI 3D Object Detection~\cite{geiger2012we}.
In total, $10165$ objects from KITTI with labels \texttt{Pedestrian}, \texttt{Cyclist}, \texttt{Car}, \texttt{Van} and \texttt{Truck} are ``cropped'' using annotated bounding boxes.

Four datasets are created with playground dimensions $K\in\{3,4,5,6\}$, and the resulting KITTI multi-object datasets are called KIMO-$K$ respectively.
Each KIMO-$K$ dataset is composed of $K\times 10000$ and $K\times 2000$ point clouds for training and testing, where each point cloud has up to $K^2$ objects.
All generated point clouds have $2048$ points, and each object in a point cloud occupies roughly the same number of points.
The KIMO-$K$ datasets are challenging, since they are composed of real LiDAR scans that are sparse and incomplete (\eg, ground-truths of the ``K'' rows in Table~\ref{tab:visual}).
For better visualization, we similarly generate datasets that we call CAD model multi-object (CAMO) which are composed of point clouds sampled from CAD models in ModelNet40~\cite{wu20153d} and ShapeNet~\cite{chang2015shapenet}.
More details are discussed in the supplementary material.

\textbf{Implementation details}:
The 2D grid $\Uv$ is defined to be $45\times 45$, and the codeword $\cv$ to be $512$-dimension.
Adam optimizer~\cite{kingma2014adam} is applied for training with a batch size $32$.
The constant $\epsilon$ in \eqref{eq:graph} is set to be 0.02.
We pre-train E-Net and F-Net for $640$ epochs with a learning rate of $2\times 10^{-4}$, then train TearingNet end-to-end for another $480$ epochs with a smaller learning rate $10^{-6}$.
Isolated points with no edges are removed from the reconstructions.
Our experiments are implemented with the PyTorch framework \cite{paszke2017automatic}.

\textbf{Benchmarks}:
We compare the TearingNet with several methods: (i)~LatentGAN~\cite{achlioptas2018learning} composed of the fully-connected layers, which exploits a much larger model than ours; (ii)~AtlasNet~\cite{groueix2018papier} with $3$ patches which has the same model size as ours; (iii)~FoldingNet~\cite{yang2018foldingnet}, and its extension, (iv) Cascaded F-Net, with two FoldingNets cascaded as $F_2(F_1(\uv;\cv); \cv)$.
Its model size is similar to ours.
Note that besides (i), the other methods all reconstruct point clouds via deforming 2D primitive(s).

We also consider several variants of the TearingNet for shape reconstruction: i)~TearingNet$_\textrm{TF}$: instead of having a trial folding first, this configuration runs a T-Net directly for topology update without considering $\Xv^{(1)}$; ii)~TearingNet$_{\overline{\textrm{GF}}}$ which excludes the graph filtering at the end; iii)~TearingNet$_{3}$ which augments the TearingNet by iterating three times, \ie, as $F\hspace{-3pt}\rightarrow\hspace{-3pt}T\hspace{-3pt}\rightarrow\hspace{-3pt}F\hspace{-3pt}\rightarrow\hspace{-3pt}T\hspace{-3pt}\rightarrow\hspace{-3pt}F$.
It is trained via loading the weights from a pre-trained TearingNet followed by a finetuning.

\def\resultpcwidth{0.35}
\def\resultgridwidth{0.13}
\def\resultzoomwidth{0.17}
\def\resultobjfactor{0.5}
\def\resultmargin{1pt}
\begin{table*}[htbp]
  \centering\scriptsize
  \caption{\small Visual comparisons of point cloud reconstructions. Points are colored according to their indices. S:~ShapeNet; T:~Torus; C:~CAMO-5; K:~KIMO-5. Objects in the red boxes are zoomed in and shown in the blue boxes.}
    \begin{tabular}{c||c|c|c|c|c}
    \hline
          & Ground-truth & AtlasNet & FoldingNet & TearingNet & Torn Grid \\
    \hline
    \hline
    S     & 
    \begin{minipage}[c]{\resultpcwidth\columnwidth}\center
        \vspace{\resultmargin}\includegraphics[width=\resultobjfactor\columnwidth]{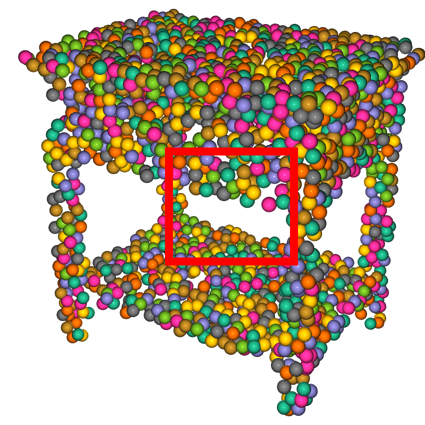}\hspace{3pt}\includegraphics[width=0.3\columnwidth]{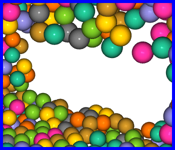}\vspace{1pt}
    \end{minipage} 
    & 
    \begin{minipage}[c]{\resultpcwidth\columnwidth}\center
        \vspace{\resultmargin}\includegraphics[width=\resultobjfactor\columnwidth]{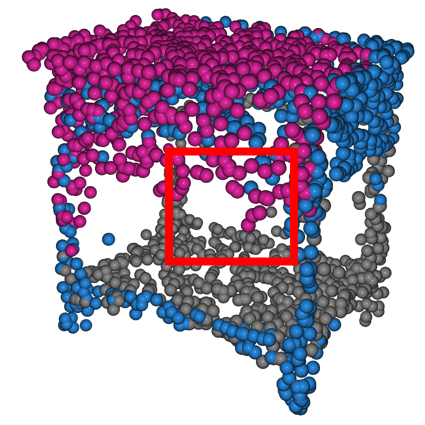}\hspace{3pt}\includegraphics[width=0.3\columnwidth]{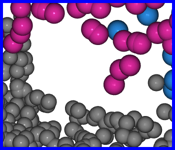}\vspace{1pt}
    \end{minipage}  
    & 
    \begin{minipage}[c]{\resultpcwidth\columnwidth}\center
        \vspace{\resultmargin}\includegraphics[width=\resultobjfactor\columnwidth]{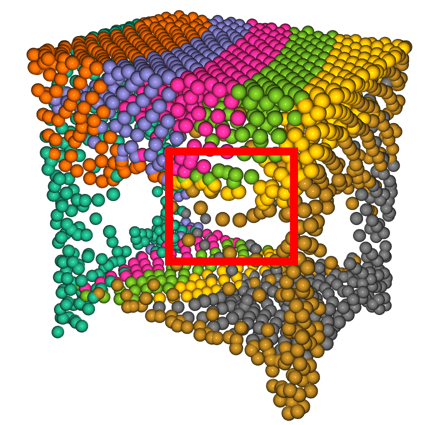}\hspace{3pt}\includegraphics[width=0.3\columnwidth]{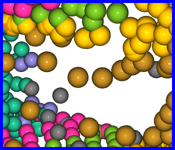}\vspace{1pt}
    \end{minipage}
    & 
    \begin{minipage}[c]{\resultpcwidth\columnwidth}\center
        \vspace{\resultmargin}\includegraphics[width=\resultobjfactor\columnwidth]{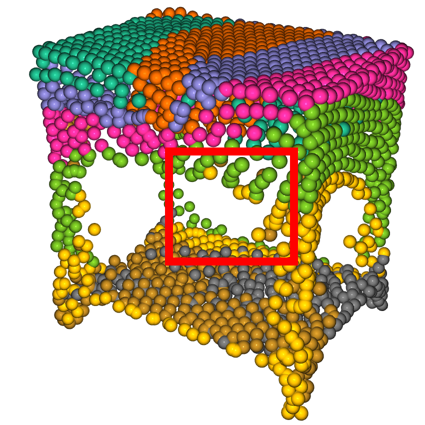}\hspace{3pt}\includegraphics[width=0.3\columnwidth]{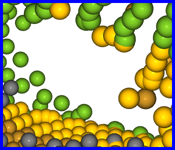}\vspace{1pt}
    \end{minipage}
    & 
    \begin{minipage}[c]{\resultgridwidth\columnwidth}\center
        \vspace{\resultmargin}\includegraphics[width=1\columnwidth]{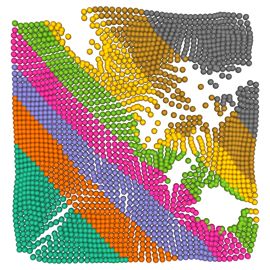}
    \end{minipage}
    \\
    \hline
    T     & 
    \begin{minipage}[c]{\resultpcwidth\columnwidth}\center
        \vspace{\resultmargin}\includegraphics[width=\resultobjfactor\columnwidth]{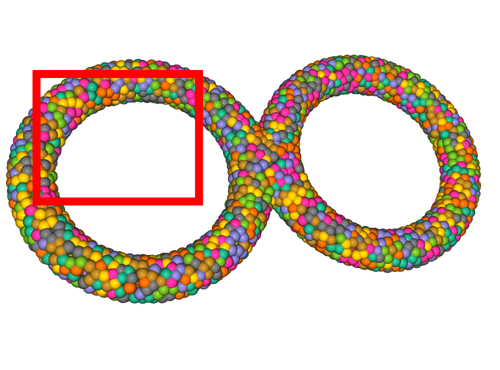}\includegraphics[width=0.3\columnwidth]{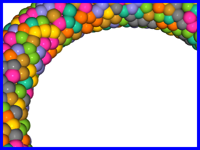}\vspace{1pt}
    \end{minipage} 
    & 
    \begin{minipage}[c]{\resultpcwidth\columnwidth}\center
        \vspace{\resultmargin}\includegraphics[width=\resultobjfactor\columnwidth]{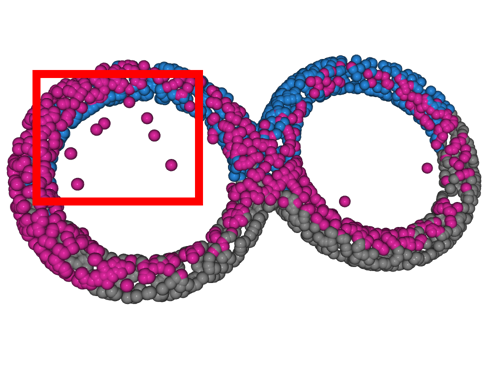}\includegraphics[width=0.3\columnwidth]{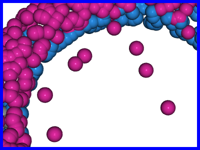}\vspace{1pt}
    \end{minipage}  
    & 
    \begin{minipage}[c]{\resultpcwidth\columnwidth}\center
        \vspace{\resultmargin}\includegraphics[width=\resultobjfactor\columnwidth]{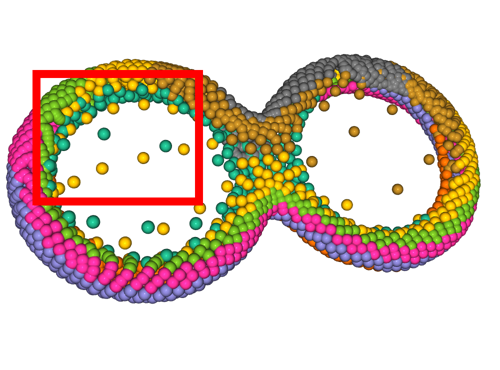}\includegraphics[width=0.3\columnwidth]{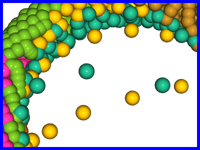}\vspace{1pt}
    \end{minipage}
    & 
    \begin{minipage}[c]{\resultpcwidth\columnwidth}\center
        \vspace{\resultmargin}\includegraphics[width=\resultobjfactor\columnwidth]{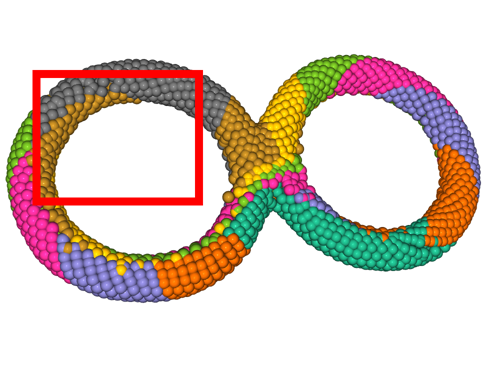}\includegraphics[width=0.3\columnwidth]{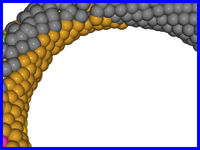}\vspace{1pt}
    \end{minipage}
    & 
    \begin{minipage}[c]{\resultgridwidth\columnwidth}\center
        \vspace{\resultmargin}\includegraphics[width=1\columnwidth]{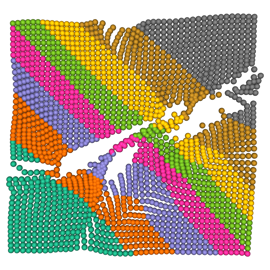}
    \end{minipage}
    \\
    \hline
    \multirow{6}[2]{*}{C} & 
    \begin{minipage}[c]{\resultpcwidth\columnwidth}\center
        \vspace{\resultmargin}\includegraphics[width=1\columnwidth]{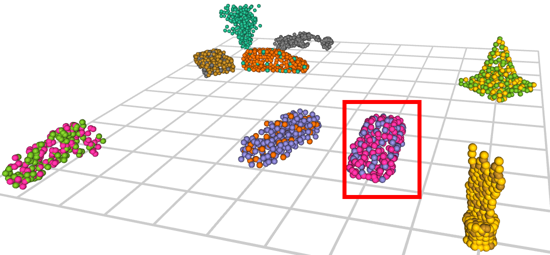}\vspace{1pt}
    \end{minipage}
    & 
    \begin{minipage}[c]{\resultpcwidth\columnwidth}\center
        \vspace{\resultmargin}\includegraphics[width=1\columnwidth]{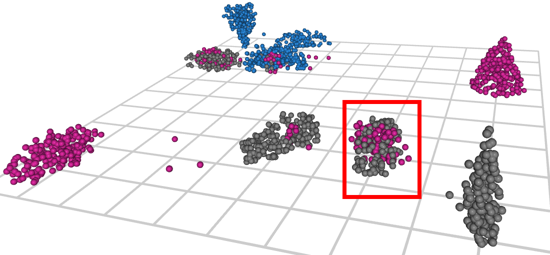}\vspace{1pt}
    \end{minipage}
    & 
    \begin{minipage}[c]{\resultpcwidth\columnwidth}\center
        \vspace{\resultmargin}\includegraphics[width=1\columnwidth]{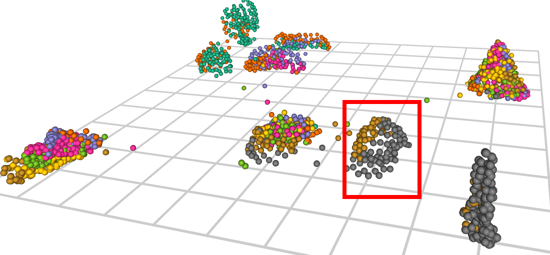}\vspace{1pt}
    \end{minipage}
    & 
    \begin{minipage}[c]{\resultpcwidth\columnwidth}\center
        \vspace{\resultmargin}\includegraphics[width=1\columnwidth]{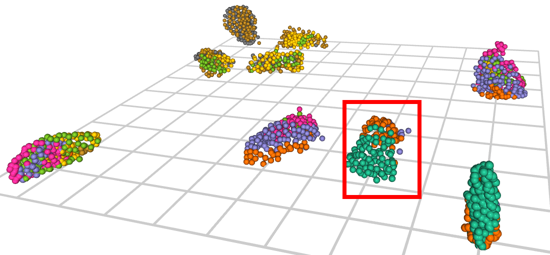}\vspace{1pt}
    \end{minipage}
    & \multirow{2}[2]{*}{
    \begin{minipage}[c]{\resultgridwidth\columnwidth}\center
        \vspace{7pt}\vspace{\resultmargin}\includegraphics[width=1\columnwidth]{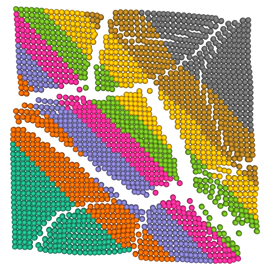}
    \end{minipage}
    } \\ & 
    \begin{minipage}[c]{\resultzoomwidth\columnwidth}\center
        \vspace{\resultmargin}\includegraphics[width=1\columnwidth]{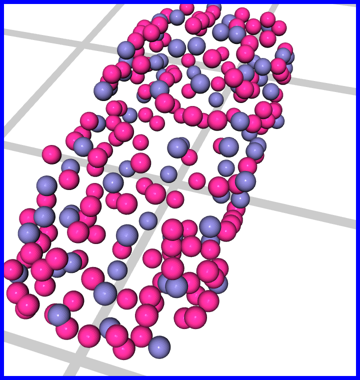}\vspace{1pt}
    \end{minipage}
    & 
    \begin{minipage}[c]{\resultzoomwidth\columnwidth}\center
        \vspace{\resultmargin}\includegraphics[width=1\columnwidth]{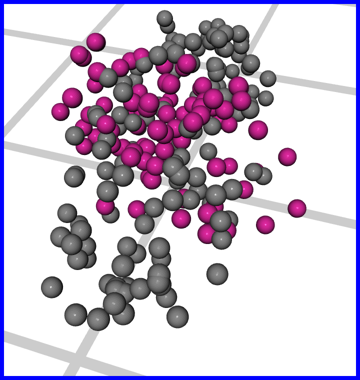}\vspace{1pt}
    \end{minipage}
    & 
    \begin{minipage}[c]{\resultzoomwidth\columnwidth}\center
        \vspace{\resultmargin}\includegraphics[width=1\columnwidth]{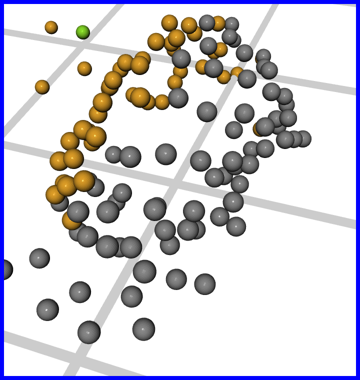}\vspace{1pt}
    \end{minipage}
    & 
    \begin{minipage}[c]{\resultzoomwidth\columnwidth}\center
        \vspace{\resultmargin}\includegraphics[width=1\columnwidth]{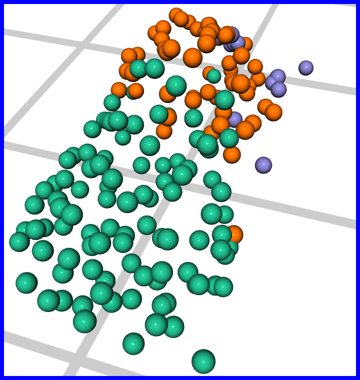}\vspace{1pt}
    \end{minipage}
    &  \\
    \hline
    \multirow{13}[4]{*}{K} & 
    \begin{minipage}[c]{\resultpcwidth\columnwidth}\center
        \vspace{\resultmargin}\includegraphics[width=1\columnwidth]{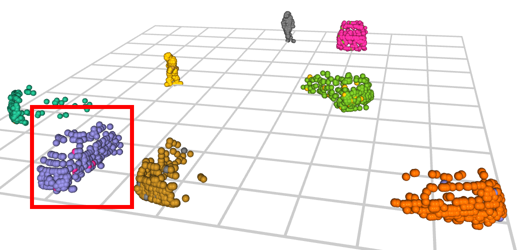}\vspace{1pt}
    \end{minipage}
    & 
    \begin{minipage}[c]{\resultpcwidth\columnwidth}\center
        \vspace{\resultmargin}\includegraphics[width=1\columnwidth]{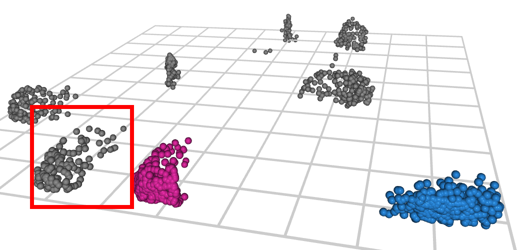}\vspace{1pt}
    \end{minipage}
    & 
    \begin{minipage}[c]{\resultpcwidth\columnwidth}\center
        \vspace{\resultmargin}\includegraphics[width=1\columnwidth]{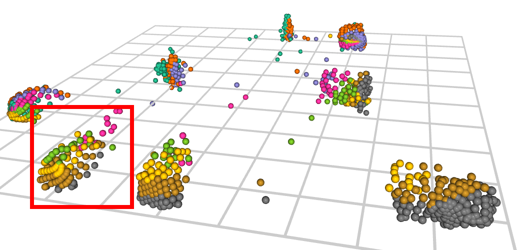}\vspace{1pt}
    \end{minipage}
    & 
    \begin{minipage}[c]{\resultpcwidth\columnwidth}\center
        \vspace{\resultmargin}\includegraphics[width=1\columnwidth]{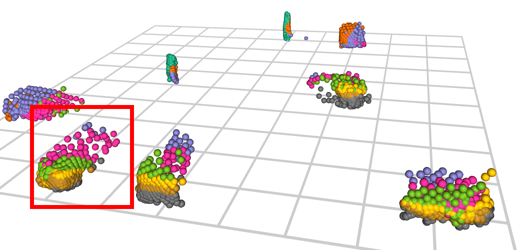}\vspace{1pt}
    \end{minipage}
    & \multirow{2}[2]{*}{
    \begin{minipage}[c]{\resultgridwidth\columnwidth}\center
        \vspace{2pt}\vspace{\resultmargin}\includegraphics[width=1\columnwidth]{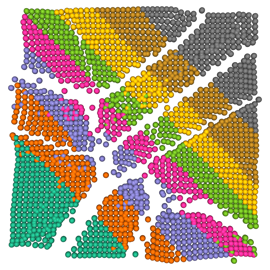}
    \end{minipage}
    } \\ & 
    \begin{minipage}[c]{\resultzoomwidth\columnwidth}\center
        \vspace{\resultmargin}\includegraphics[width=1\columnwidth]{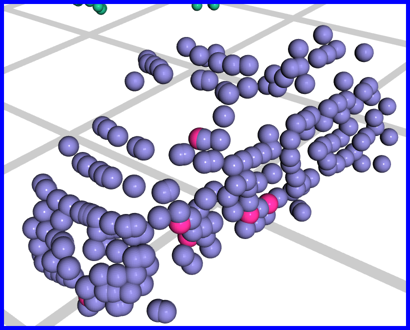}\vspace{1pt}
    \end{minipage}
    & 
    \begin{minipage}[c]{\resultzoomwidth\columnwidth}\center
        \vspace{\resultmargin}\includegraphics[width=1\columnwidth]{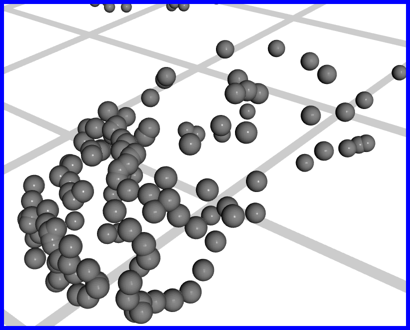}\vspace{1pt}
    \end{minipage}
    & 
    \begin{minipage}[c]{\resultzoomwidth\columnwidth}\center
        \vspace{\resultmargin}\includegraphics[width=1\columnwidth]{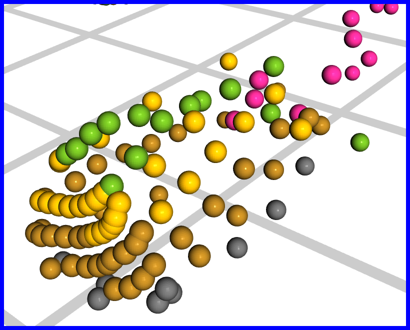}\vspace{1pt}
    \end{minipage}
    & 
    \begin{minipage}[c]{\resultzoomwidth\columnwidth}\center
        \vspace{\resultmargin}\includegraphics[width=1\columnwidth]{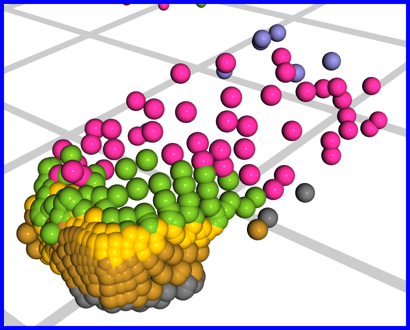}\vspace{1pt}
    \end{minipage}
    &  \\
    \cline{2-6} & 
    \begin{minipage}[c]{\resultpcwidth\columnwidth}\center
        \vspace{\resultmargin}\includegraphics[width=1\columnwidth]{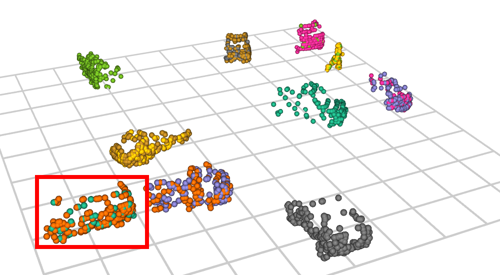}\vspace{1pt}
    \end{minipage}
    & 
    \begin{minipage}[c]{\resultpcwidth\columnwidth}\center
        \vspace{\resultmargin}\includegraphics[width=1\columnwidth]{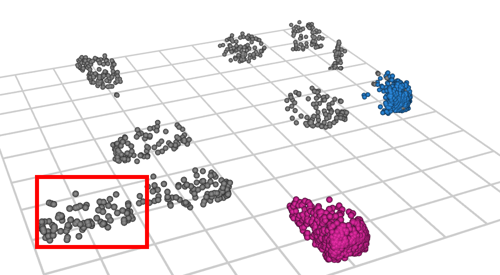}\vspace{1pt}
    \end{minipage}
    & 
    \begin{minipage}[c]{\resultpcwidth\columnwidth}\center
        \vspace{\resultmargin}\includegraphics[width=1\columnwidth]{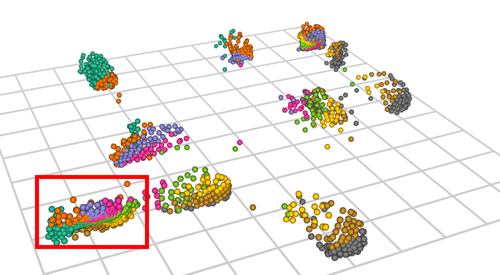}\vspace{1pt}
    \end{minipage}
    & 
    \begin{minipage}[c]{\resultpcwidth\columnwidth}\center
        \vspace{\resultmargin}\includegraphics[width=1\columnwidth]{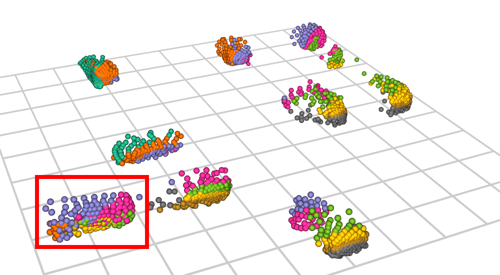}\vspace{1pt}
    \end{minipage}
    & \multirow{2}[2]{*}{
    \begin{minipage}[c]{\resultgridwidth\columnwidth}\center
        \vspace{\resultmargin}\includegraphics[width=1\columnwidth]{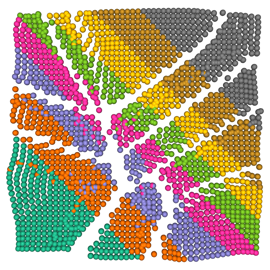}
    \end{minipage}
    } \\ & 
    \begin{minipage}[c]{\resultzoomwidth\columnwidth}\center
        \vspace{\resultmargin}\includegraphics[width=1\columnwidth]{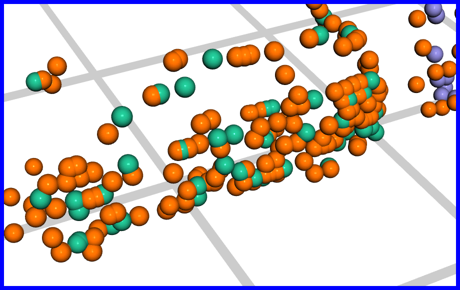}\vspace{1pt}
    \end{minipage}
    & 
    \begin{minipage}[c]{\resultzoomwidth\columnwidth}\center
        \vspace{\resultmargin}\includegraphics[width=1\columnwidth]{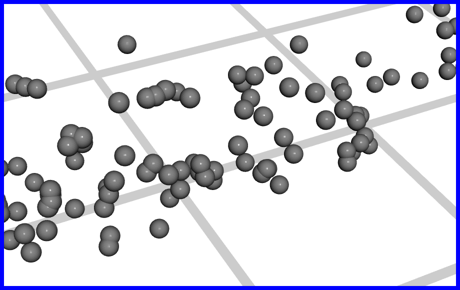}\vspace{1pt}
    \end{minipage}
    & 
    \begin{minipage}[c]{\resultzoomwidth\columnwidth}\center
        \vspace{\resultmargin}\includegraphics[width=1\columnwidth]{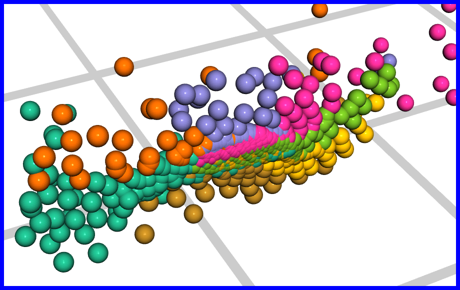}\vspace{1pt}
    \end{minipage}
    & 
    \begin{minipage}[c]{\resultzoomwidth\columnwidth}\center
        \vspace{\resultmargin}\includegraphics[width=1\columnwidth]{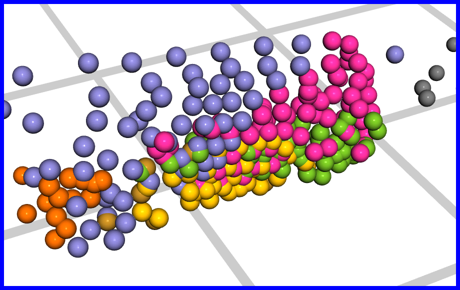}\vspace{1pt}
    \end{minipage}
    &  \\
    \hline
    \end{tabular}%
  \label{tab:visual}%
\end{table*}%

\begin{table*}[htbp]
  \centering\scriptsize
  \caption{\small Evaluation of 3D point cloud reconstruction, in terms of both CD and EMD.}
    \begin{tabular}{c||cc|cccc||cc|cccc}
    \hline
    Metrics & \multicolumn{6}{c||}{CD ($\times 10^{-2}$)} & \multicolumn{6}{c}{EMD}\\
    \hline
    Datasets & ShapeNet & Torus & KI.-3 & KI.-4 & KI.-5 & KI.-6 & ShapeNet & Torus & KI.-3 & KI.-4 & KI.-5 & KI.-6 \\
    \hline
    \hline
    LatentGAN & 2.85  & 2.45  & 7.10  & 11.64 & 17.18 & 19.21 & 0.218 & 0.202 & 1.982 & 3.231 & 3.773 & 4.721 \\
    AtlasNet & 2.72  & 2.41  & \textbf{4.50} & 6.76  & 9.59  & 11.63 & \textbf{0.163} & \textbf{0.146} & 1.333 & 2.811 & 3.173 & 4.440 \\
    FoldingNet & 2.75  & 1.90  & \textbf{4.72} & 6.57  & 9.01  & 11.06 & 0.372 & 0.191 & 1.748 & 2.864 & 3.056 & 4.569 \\
    Cascaded F-Net & 2.69  & 1.95  & 4.77  & 6.67  & 9.13  & 10.94 & 0.207 & 0.196 & 1.533 & 2.381 & 2.944 & 4.189 \\
    \hline
    TearingNet$_{\textrm{TF}}$ & 2.59  & 1.84  & 5.05  & 6.58  & 8.55  & 10.86 & 0.206 & 0.163 & 1.055 & 1.787 & 2.565 & 3.476 \\
    TearingNet$_{\overline{\textrm{GF}}}$ & 2.56  & 1.74  & 4.92  & 6.45  & 8.29  & 10.29 & 0.172 & 0.170 & 0.958 & 1.441 & 1.879 & 2.648 \\
    TearingNet (Ours) & \textbf{2.54} & \textbf{1.72} & 4.78  & \textbf{6.43} & \textbf{8.29} & \textbf{10.23} & 0.174 & 0.156 & \textbf{0.940} & \textbf{1.438} & \textbf{1.872} & \textbf{2.614} \\
    TearingNet$_{\text{3}}$ (Ours) & \textbf{2.53} & \textbf{1.73} & 4.74  & \textbf{6.42} & \textbf{8.24} & \textbf{10.15} & \textbf{0.169} & \textbf{0.143} & \textbf{0.941} & \textbf{1.361} & \textbf{1.867} & \textbf{2.315} \\
    \hline
    \end{tabular}%
  \label{tab:rec_kimo}%
  \vspace{-10pt}
\end{table*}%

\subsection{Performance Comparison}\label{ssec:comp}
\textbf{Reconstruction}: We first evaluate the reconstruction quality of the proposed TearingNet.
Table~\ref{tab:visual} visualizes some reconstructed point clouds from several datasets. 
Compared to the TearingNet, FoldingNet leaves more ``ghost'' points outside object surfaces, while AtlasNet results in unbalanced point distributions.
In contrast, TearingNet produces point clouds that look \emph{clean} and \emph{orderly}, with appearances close to the inputs.
The 2D-grids (last column of Table~\ref{tab:visual}) are torn apart to accommodate the corresponding 3D topologies.
It demonstrates how object topologies are \emph{discovered} and \emph{utilized} via the TearingNet architecture.

We report the augmented Chamfer Distance (referred to as CD) and the Earth Mover's Distance (EMD) \cite{fan2017point} of the competing methods in Table~\ref{tab:rec_kimo}.
For both metrics, a smaller number indicates more accurate reconstruction.
In general, our TearingNet and TearingNet$_3$ outperform the benchmarks, which is even more obvious in terms of EMD.
As the topology complexity grows from KIMO-3 to KIMO-6, our method outperforms the competitors more significantly.
By comparing our TearingNet to TearingNet$_{\textrm{TF}}$, we see the effectiveness of inserting a Folding network before the Tearing network.
That is because, with the first trial folding result, the Tearing network can better capture the discrepancy between a genus zero topology and the ground-truth topology via back-propagation.
From Table~\ref{tab:rec_kimo}, we also see that, begin with the TearingNet$_{\overline{\textrm{GF}}}$, by first incorporating the graph filter (TearingNet), then further iterating the T-Net and the F-Net (TearingNet$_3$), reconstruction qualities continues to improve.
We observe similar results on the CAMO datasets.

We also experimented with recent methods AtlasNetV2~\cite{deprelle2019learning} and 3D Point Capsule Network~\cite{zhao20193d} and observe good performance on single-object datasets.
On ShapeNet, AtlasNetV2 has even achieved a lowest CD of $2.48\times 10^{-2}$.
However, both methods fail to converge on our multi-object datasets.
We conjecture that is because both of them are over-optimized for single-object cases, \eg, AtlasNetV2 specifically learns the elementary structure of objects.
Hence, the diversified scene configurations in our multi-object datasets ``confuse'' these two methods.

{\bf Object counting}:
In a multi-object scene, adding objects yields a more complex topology.
From the multi-object examples in Table~\ref{tab:visual}, we see that the number of torn patches in a 2D-grid approximately equals the number of objects in a scene.
It implies that the latent codewords from TearingNet are \emph{aware of the geometric topology}.
To further affirm their representativeness of topologies, we next try to ``count'' the object number directly from the codewords.
In practice, counting is a critical task in applications such as traffic jam detection and crowd analysis \cite{lempitsky2010learning,onoro2016towards}.

In this task, the TearingNet and other benchmarks trained from the reconstruction experiment are carried over.
We again use the challenging KIMO datasets to experiment with this use case.
As a preparation, we feed the test dataset to the PointNet encoder to collect codewords.
Next, we employ 4-fold cross-validation to train/test an SVM classifier: codewords are equally divided into 4 folds, then \emph{only one} of the four is used to train the SVM together with their count labels while the other three are reserved for counting test.
SVM is selected for the test as it would not modify the feature space learned by autoencoders.
Consequently, this setting overall requires a small number of labels, because our feature (codeword) learning is achieved in an \emph{unsupervised} manner, while the counting task is learned in a \emph{weakly supervised} manner.

The counting performance is measured by mean absolute error (MAE) between the predicted counting and the ground-truth counting~\cite{zhang2015cross}.
As shown in the upper-half of Table~\ref{tab:cnt_det}, TearingNet and TearingNet$_3$ consistently produce the smallest MAEs.
For example, on KIMO-4, TearingNet brings down MAE by more than 40\% comparing to FoldingNet/AtlasNet, showing its strong capability in representing scene topologies.

\begin{figure}[t]
  \centering\scriptsize
  \subfloat[t-SNE visualization]{
  \includegraphics[width=.35\columnwidth]{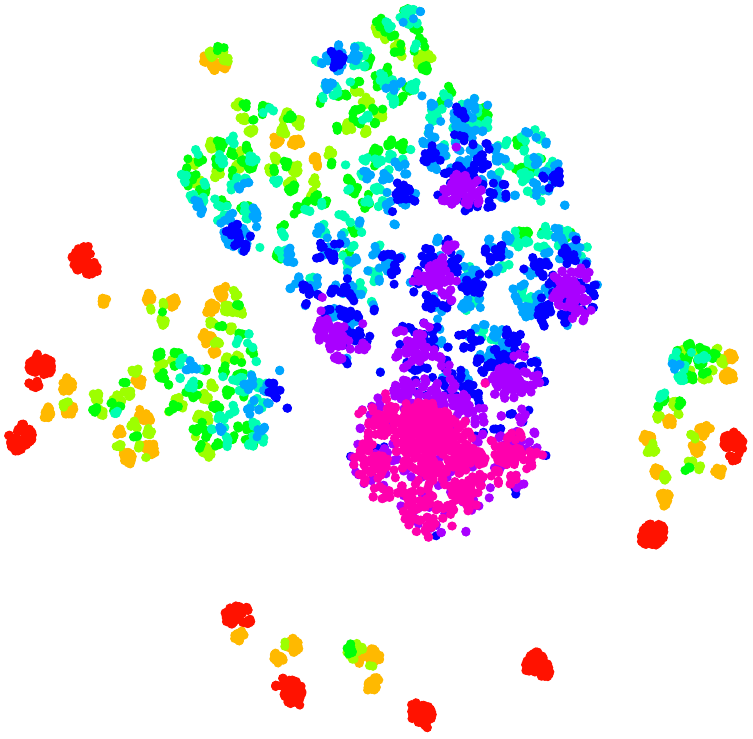}\hspace{3pt}
  \includegraphics[width=.1\columnwidth]{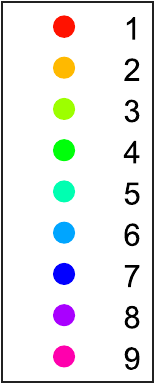}\label{fig:tsne_cluster}}\hspace{12pt}
  \subfloat[$d_k$ vs. object count $k$]{\includegraphics[width=.35\columnwidth]{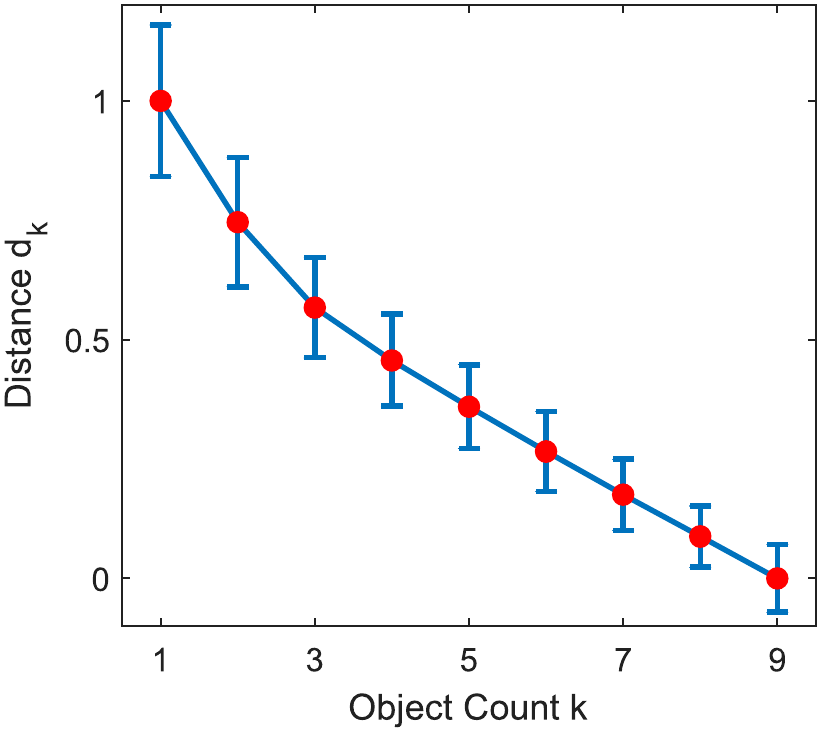}\label{fig:tsne_distance}}
  \caption{\small Analyzing the feature space of the TearingNet.}
  \label{fig:tsne}
  \vspace{-10pt}
\end{figure}

We further inspect the feature space learned by the TearingNet to understand how it is linked to the topology.
We first collect the TearingNet codewords of the KIMO-3 dataset and visualize them using t-SNE, as shown in Figure~\ref{fig:tsne_cluster}.
Here the points are colored based on their corresponding counting labels.
Note that for the $3\times 3$ playground in KIMO-3, there are 9 and 36 combinations when placing 1 and 2 objects, respectively.
Correspondingly, 9 and 36 clusters could be observed in the t-SNE figure.
And as there is only 1 possible combination to arrange 9 objects, all points representing 9 objects aggregate to a single cluster.

\begin{table}[t]
  \centering\scriptsize
  \caption{\small Evaluation of object counting and object detection.}
    \begin{tabular}{ c||c|cccc}
    \hline
    \multirow{2}[3]{*}{Tasks} & \multirow{2}[3]{*}{Methods} & \multicolumn{4}{c}{Datasets} \\
\cline{3-6}          &       & KI.-3 & KI.-4 & KI.-5 & KI.-6 \\
    \hline
    \hline
    \multicolumn{1}{c||}{\multirow{8}[3]{*}{
    \begin{minipage}[t]{0.17\columnwidth}
    \hspace{9pt}Counting\\
    (MAE,\,$\times 10^{-1}$)
    \end{minipage}
    }} & LatentGAN & 0.671 & 8.439 & 14.079 & 14.523 \\
          & AtlasNet & 0.125 & 2.908 & 6.727 & 8.569 \\
          & FoldingNet & 0.204 & 3.031 & 6.340 & 8.527 \\
          & Cascaded F-Net & 0.270 & 3.207 & 7.104 & 9.792 \\
\cline{2-6}          & TearingNet$_{\textrm{TF}}$ & 0.127 & 2.207 & 5.716 & 8.346 \bigstrut[t]\\
          & TearingNet$_{\overline{\textrm{GF}}}$ & 0.123 & 1.805 & 5.105 & 8.044 \\
          & \hspace{-2pt}TearingNet\,(Ours)\hspace{-2pt} & \textbf{0.123} & \textbf{1.721} & \textbf{5.079} & \textbf{8.026} \\
          & \hspace{-5pt}TearingNet$_3$\,(Ours)\hspace{-5pt} & \textbf{0.121} & \textbf{1.740} & \textbf{5.050} & \textbf{7.992} \\
    \hline
    \hline
    \multicolumn{1}{c||}{\multirow{8}[3]{*}{
    \begin{minipage}[t]{0.17\columnwidth}
    \hspace{5pt}Detection\\
    (Accuracy, \%)
    \end{minipage}
    }} & LatentGAN & \textbf{93.53} & 63.78 & 65.65 & 78.80 \\
          & AtlasNet & 88.84 & 73.79 & 73.58 & 83.53 \\
          & FoldingNet & 92.71 & 80.12 & 77.10 & 82.92 \\
          & Cascaded F-Net & 89.78 & 76.36 & 76.84 & 82.63 \\
\cline{2-6}          & TearingNet$_{\textrm{TF}}$ & 93.33 & 82.83 & 78.43 & 83.74 \bigstrut[t]\\
          & TearingNet$_{\overline{\textrm{GF}}}$ & 93.44 & 83.42 & 79.70 & 84.55 \\
          & \hspace{-2pt}TearingNet\,(Ours)\hspace{-2pt} & 93.42 & \textbf{83.42} & \textbf{79.74} & \textbf{84.55} \\
          & \hspace{-5pt}TearingNet$_3$\,(Ours)\hspace{-5pt} & \textbf{93.46} & \textbf{83.44} & \textbf{79.72} & \textbf{86.52} \\
    \hline
    \end{tabular}%
  \label{tab:cnt_det}%
  \vspace{-14pt}
\end{table}%

Finally, the appearance of the t-SNE diagram exhibits a \emph{tree} structure. 
When inspecting one cluster of a larger counting (\eg, 9, 8, \etc), it is always surrounded by several smaller counting clusters (\eg, 8, 7, \etc).
This observation is actually due to a recursive encapsulation from counting 1 to 9 where counting 9 stays at the center. 
If we compute an average Euclidean distance $d_k$ from all codewords of counting $k$ to the mean codeword of counting 9, we observe that $d_k$ approximately linearly increases as object counting $k$ decreases (see Figure~\ref{fig:tsne_distance}, where the distances are normalized to $[0,1]$, the error bars are also shown). 
It implies that the codewords distribute in a layered manner with respect to counting (\ie, topology) and they are \emph{topology-aware}.

{\bf Object detection}:
Having revealed the superiority of our proposal in point reconstruction and topology understanding, we finally devised a last experiment to demonstrate such superiority in low/mid-level tasks can be transferred to high-level understanding tasks.
Specially, we consider the pedestrian detection task which is critical under an autonomous driving scenario~\cite{dollar2009pedestrian}.
Similar to object counting, we train binary SVM classifiers and evaluate their performance using a $4$-fold cross-validation strategy.
Again, one fold is applied for training and the rest for testing.
Detection accuracy is collected in the bottom-half of Table~\ref{tab:cnt_det}.
Note that KIMO-3 is the easiest dataset as it contains the least combination possibilities, and thus the simple LatentGAN already provides good accuracy.
Compared to the other benchmarks, both TearingNet and TearingNet$_3$ perform comparable on KIMO-3 and significantly better on KIMO-4, -5, and -6.
Moreover, for KIMO-4, TearingNet surpasses AtlasNet and FoldingNet by about 10\% and 3\%, respectively.

%% file: 6_conclude.tex
We consider the problem of representing and reconstructing point clouds of ample topologies with an autoencoder, given the latent representations in the form of a fixed-length vector.
To tackle this task, we propose a TearingNet architecture which iteratively discovers and utilizes topology with a Tearing network and a Folding network, respectively.
The superior capability of our proposal is demonstrated in terms of shape reconstruction and producing topology-friendly representations for point clouds.
Essentially, the Tearing network reparameterizes the surface defined by the Folding network according to a learned topology.
For future research, we plan to apply the TearingNet for scene point clouds with natural object placement.
We are also interested in applying its generalization---the GCAE---to other data modalities where topology matters, such as images and videos.
